\documentclass[11pt,a4paper]{article}

\usepackage[T1]{fontenc}
\usepackage[utf8]{inputenc}
\IfFileExists{lmodern.sty}{\usepackage{lmodern}}{\usepackage{mathptmx}}
\usepackage{microtype}
\usepackage[margin=2.5cm]{geometry}
\usepackage{amsmath}
\usepackage{graphicx}
\usepackage{booktabs}
\usepackage{tabularx}
\usepackage{array}
\usepackage{xcolor}
\usepackage{tikz}
\usetikzlibrary{positioning,arrows.meta}
\usepackage[numbers,sort&compress]{natbib}
\usepackage[hidelinks]{hyperref}
\usepackage[capitalise,noabbrev]{cleveref}
\usepackage{authblk}

\definecolor{ecblue}{HTML}{0072B2}
\definecolor{ecgreen}{HTML}{009E73}
\definecolor{ecorange}{HTML}{E69F00}

\newcommand{\ER}{\textsc{EasyResearch}}
\newcommand{\EC}{\textsc{EasyClassifier}}
\newcommand{\EV}{\textsc{EasyVision}}

\newcommand{\TBDDOI}{\href{https://doi.org/10.5281/zenodo.23305077}{doi:10.5281/zenodo.23305077} (all versions)}

\newcommand{\RegNData}{9}
\newcommand{\RegRuns}{90}
\newcommand{\RegSeeds}{10}
\newcommand{\RegBiasNaive}{+2.2}
\newcommand{\RegBiasComp}{+2.5}
\newcommand{\RegBiasFinal}{$-$1.0}
\newcommand{\RegMAENaive}{8.2}
\newcommand{\RegMAEComp}{7.9}
\newcommand{\RegMAEFinal}{9.1}
\newcommand{\RegP}{1.3\times10^{-4}}
\newcommand{\RegPds}{0.0098}

\newcommand{\RegCtrlCompVal}{$-$0.043}
\newcommand{\RegCtrlFinalVal}{$-$0.089}
\newcommand{\RegCtrlTruthVal}{$-$0.157}
\newcommand{\ScrOut}{0.27}
\newcommand{\ScrIn}{$-$0.33}
\newcommand{\ScrExt}{$-$0.44}
\newcommand{\FcNData}{15}
\newcommand{\FcShufGeo}{19}
\newcommand{\FcShufMedian}{25}

\newcommand{\FcFinalGeo}{1.5}
\newcommand{\FcShufTyp}{1.48}
\newcommand{\FcCompTyp}{1.28}
\newcommand{\FcFinalTyp}{1.46}
\newcommand{\FcShufBelow}{9}

\newcommand{\FcP}{0.1}
\newcommand{\FcToolBetter}{7}
\newcommand{\FcSame}{5}
\newcommand{\FcToolWorse}{3}
\newcommand{\FcTruthGain}{14}
\newcommand{\FcBeatsRule}{12}
\newcommand{\FcBeatsRuleShuf}{10}
\newcommand{\FcRWn}{5}
\newcommand{\FcRWShufML}{5}
\newcommand{\FcRWShufWorse}{5}
\newcommand{\FcRWShufGain}{20}
\newcommand{\FcRWToolNaive}{4}
\newcommand{\SigNData}{4}
\newcommand{\SigRuns}{12}
\newcommand{\SigSeeds}{3}
\newcommand{\SigBiasComp}{0.00}
\newcommand{\SigBiasFinal}{$-$0.02}
\newcommand{\SigMAEComp}{1.9}
\newcommand{\SigMAEFinal}{1.8}

\newcommand{\SigCtrlComp}{46.9}
\newcommand{\SigCtrlFinal}{42.7}
\newcommand{\SigCtrlTruth}{52.9}

\title{\ER: Honest Regression, Forecasting and Signal Classification\\ for Researchers Who Do Not Program}

\author[1]{Ahmad B. A. Hassanat\thanks{Corresponding author: \href{mailto:ahmad.hassanat@gmail.com}{ahmad.hassanat@gmail.com}}}
\author[2]{Ghada A. Altarawneh\thanks{\href{mailto:ghadaa@mutah.edu.jo}{ghadaa@mutah.edu.jo}}}
\affil[1]{Faculty of Information Technology, Mutah University, Karak, Jordan}
\affil[2]{Department of Accounting, Mutah University, Karak, Jordan}
\date{}

\begin{document}
\maketitle

\begin{abstract}
\EC{} showed that researchers who do not program can obtain honest classification results when the correct evaluation is the default. Most applied studies, however, predict a number, forecast a series or classify recordings, and each task has its own way of producing scores that do not hold on new data. We present \ER, open-source analysis cores for regression, time-series forecasting and signal classification, and \ER{} Desktop, a Windows application that runs these three tasks and \EC's classification without code. All four follow the same rules: everything learned from data is learned on training data only, models keep their published defaults with fixed seeds, the selected model receives a separate final score, forecasts are never evaluated on shuffled data, and every run writes a report with a Methods paragraph and the references for every method used. We measured whether the reported scores are honest by keeping part of each dataset untouched. In regression (\RegNData{} public datasets, \RegRuns{} runs), the common practice overstated $R^2$ by \RegBiasNaive{} points on average and the best comparison score by \RegBiasComp, while \ER's final score was \RegBiasFinal{} (slightly conservative); screening predictors on all rows before cross-validation, which \ER{} does not offer, reported $R^2 = \ScrOut$ on pure noise against \ScrExt{} on new data. In forecasting (\FcNData{} series), shuffled cross-validation over windows of past values promised errors \FcShufGeo\% smaller than its chosen models achieved in the untouched future (geometric mean; median \FcShufMedian\%), while \ER's final-period score was unbiased to within \FcFinalGeo\%; the models \ER{} chose had \FcTruthGain\% lower future error, and on random walks shuffled cross-validation chose a machine-learning model in \FcRWShufML{} of \FcRWn{} series, each worse in the future than repeating the last value, which \ER{} chose \FcRWToolNaive{} times. In signal classification (\SigNData{} UCR datasets) one method won every comparison and neither estimate was biased. The software is available under the MIT license.
\end{abstract}


\begin{table}[h]
\centering\small
\caption{Code metadata.}
\label{tab:metadata}
\begin{tabularx}{\linewidth}{@{}lX@{}}
\toprule
Versions & \ER{} 0.1.2 (analysis cores); \ER{} Desktop 0.1.2; \EC{} 0.8.2 \\
Code repository & \url{https://github.com/ahmadhassant/easyresearch} (desktop application in \texttt{desktop}) \\
Archived version (DOI) & \TBDDOI \\
License & MIT \\
Language and requirements & Python $\geq$ 3.10 with pandas, NumPy, scikit-learn, statsmodels, PyTorch, matplotlib; desktop: C\#/.NET (WPF), Windows x64, bundled Python runtime. The benchmarks in \cref{sec:evaluation} were run with \ER{} 0.1.0; versions 0.1.1 and 0.1.2 changed no benchmarked analysis \\
Documentation & README files of both folders; protocol of the desktop worker (\texttt{desktop/docs/PROTOCOL.md}) \\
Support & GitHub issues; \href{mailto:ahmad.hassanat@gmail.com}{ahmad.hassanat@gmail.com} \\
\bottomrule
\end{tabularx}
\end{table}

\section{Motivation and significance}
\label{sec:motivation}

Leakage and optimistic evaluation are among the most common reasons why published machine-learning results fail to replicate \citep{kapoor2023leakage,lones2021pitfalls}. For classification of tabular data we built \EC{} \citep{hassanat2026easyclassifierpaper}, a guided tool that makes the honest procedure the default; on public data from seven fields, the common practice of fitting preprocessing on all rows and reporting the winner's cross-validation score overstated balanced accuracy by 2.9 percentage points on average, against 0.7 for \EC's reported score.

Applied researchers ask other questions as often. A clinician predicts a laboratory value, an economist a wage, an engineer a load; a hydrologist forecasts river flow, a retailer next year's sales; a cardiologist classifies heartbeats, a physiotherapist movement recordings. The same two errors appear in all of them --- learning from the test data before splitting, and reporting the score of the model that won a comparison \citep{varma2006bias,cawley2010overfitting} --- and each task adds its own:
\begin{itemize}
  \item \emph{Regression} invites screening the predictors by their correlation with the outcome on all rows and cross-validating only the final model, the ``wrong way'' to cross-validate \citep{hastie2009esl}.
  \item \emph{Forecasting} has a direction in time. Shuffled cross-validation over windows of past values lets a model learn from values later than those it predicts \citep{tashman2000,bergmeir2012cv}. It can be valid for purely autoregressive models with uncorrelated errors \citep{bergmeir2018note}, but the user of a no-code tool cannot check that condition, and the evaluation of statistical models has no shuffled equivalent at all.
  \item \emph{Signal classification} uses expensive feature extractors and deep networks whose comparison invites reporting the best of several runs.
\end{itemize}

\ER{} extends \EC's answer to these three tasks and puts all four into one desktop application. This paper describes the design (\cref{sec:design}), the three cores (\cref{sec:cores}), the desktop application (\cref{sec:desktop}) and one example of each task (\cref{sec:examples}), and then measures, on public data with an untouched reference part, whether the scores the software reports are honest (\cref{sec:evaluation}).

\section{Design}
\label{sec:design}

The cores share \EC's five principles \citep{hassanat2026easyclassifierpaper}: no code and no jargon unless wanted; honest by default; default parameters without tuning, since tuning adds a further layer of selection and transfers poorly from one small dataset \citep{probst2019tunability}; exact reproducibility with fixed seeds; and output that is ready to report. Two rules are added for the new tasks:

\paragraph{Time is never shuffled.} A series is split once, in time order: a development part and a final period (about the last 20\%). Models are compared at up to five rolling forecast origins in the development part, each model retrained on the values before each origin \citep{tashman2000}; the selected model is then scored in the final period, again retrained before each origin. Only horizons the series can evaluate in this way are offered. An automated test checks that no model's forecast changes when every value after the forecast origin is changed.

\paragraph{Every method is cited.} Each report's Methods paragraph cites every model, distance, evaluation design and measure that was used, and \texttt{citations.txt} lists exactly the report's references with what each is cited for.

\begin{figure}[t]
\centering
\resizebox{\linewidth}{!}{%
\begin{tikzpicture}[
  node distance=5mm and 4mm,
  box/.style={draw=ecblue, very thick, rounded corners=2pt, fill=ecblue!6,
              align=center, font=\small, text width=27mm, minimum height=13mm},
  core/.style={draw=ecgreen, very thick, rounded corners=2pt, fill=ecgreen!8,
               align=center, font=\small, text width=34mm, minimum height=13mm},
  arr/.style={-{Stealth[length=2.2mm]}, thick, black!70}]
\node[box] (gui) {\ER{} Desktop\\ \scriptsize WPF, C\#};
\node[box, right=of gui] (proto) {Worker process\\ \scriptsize JSON-lines protocol};
\node[core, right=of proto, yshift=12mm] (cls) {Classification\\ \scriptsize \EC{} 0.8.2};
\node[core, right=of proto, yshift=-12mm] (reg) {Regression\\ \scriptsize \texttt{easyresearch.regression}};
\node[core, right=of cls] (fc) {Forecasting\\ \scriptsize \texttt{easyresearch.timeseries}};
\node[core, right=of reg] (sig) {Signals\\ \scriptsize \texttt{easyresearch.timeseries}};
\node[box, below=10mm of proto, text width=30mm] (cli) {Command line\\ \scriptsize \texttt{easyresearch \ldots}};
\draw[arr] (gui) -- (proto);
\draw[arr] (proto.east) -- (cls.west);
\draw[arr] (proto.east) -- (reg.west);
\draw[arr] (proto.east) -- ++(4mm,0) |- (fc.west);
\draw[arr] (proto.east) -- ++(4mm,0) |- (sig.west);
\draw[arr] (cli.east) -| ([xshift=-2mm]reg.south);
\end{tikzpicture}}
\caption{Architecture. The desktop application starts a separate Python worker for every job and talks to it through a versioned JSON-lines protocol; the worker runs the same cores as the command line. Each task is a module with a manifest and an adapter, so a new task (for example multichannel signals) needs no change to the others.}
\label{fig:architecture}
\end{figure}
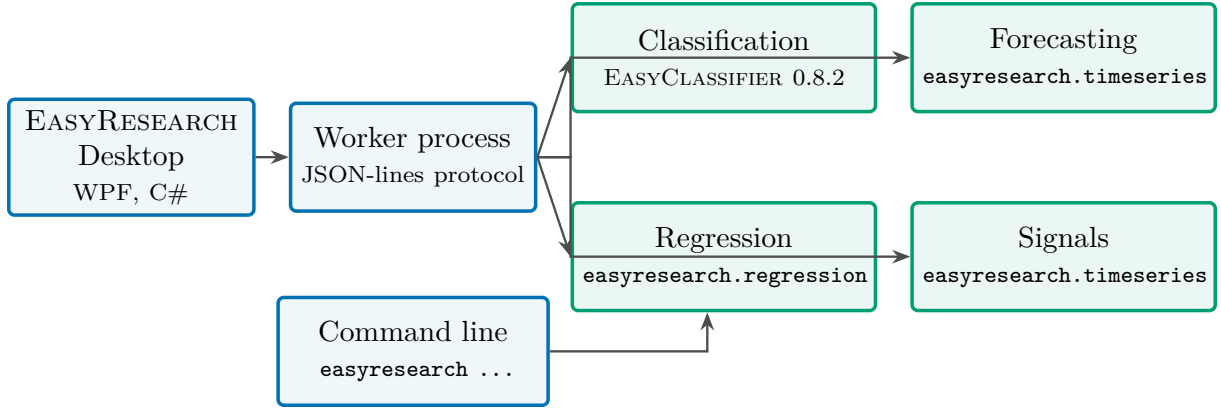

\section{The analysis cores}
\label{sec:cores}

\subsection{Regression}
Regression reuses \EC's data loading and leakage-free preprocessing \citep{pedregosa2011sklearn}. Ten models are offered: linear and ridge regression, decision tree, random forest \citep{breiman2001rf}, histogram gradient boosting, support vector regression, KNN with the Hassanat distance \citep{hassanat2014arxiv,hassanat2014distance} (columns scaled to 0--1 on training rows; the sign-symmetric form \citep{hassanat2014distance2} for negative values), a neural network, and XGBoost and LightGBM when installed; the fast default set is linear regression, random forest and gradient boosting. SVR and the neural network learn on a standardised target, fitted on training rows, because their defaults assume values of roughly unit size; this is a change of units, not tuning. Models are ranked by $R^2$ on held-out rows, and the selected model receives a final score by nested cross-validation (up to 2{,}000 rows) or on an untouched 20\% test set. Every result is shown next to the reference of always predicting the training mean. Measures are $R^2$, RMSE, MAE, median absolute error, MAPE (only when no actual value is zero or changes sign) and the mean Hassanat similarity percentage, MHSP \citep{hassanat2024mhsp}.

\subsection{Forecasting}
One series with a date column (hourly to yearly; the season length follows) or in row order. Reference rules (naive, seasonal naive) are always included, because a model is useful only if it beats them \citep{hyndman2021fpp3}. Statistical models are exponential smoothing with damped trend and seasonality \citep{hyndman2002ets} and the Theta method \citep{assimakopoulos2000theta}, from statsmodels \citep{seabold2010statsmodels}. Machine-learning models (ridge, random forest, gradient boosting, KNN with the Hassanat distance) predict the next $H$ values at once from the last $L$ values \citep{bontempi2013strategies}, with windows expressed relative to their last value so that tree and neighbour models follow trends. Deep networks (MLP, LSTM, GRU, TCN, N-BEATS, Transformer; PyTorch \citep{paszke2019pytorch}) are optional, because each takes minutes. The model is chosen by MASE \citep{hyndman2006mase} by default; MAE, RMSE, sMAPE and MHSP are reported, with an 80\% band from the comparison errors whose coverage in the final period is checked.

\subsection{Signal classification}
One recording of a single channel per row, from a table or a UCR archive file \citep{dau2019ucr}. Models are KNN with the Hassanat distance, ROCKET with 10{,}000 random convolution kernels \citep{dempster2020rocket}, signal features with a random forest \citep{fulcher2014features}, and optional deep networks (FCN, ResNet, InceptionTime, LSTM). Evaluation is \EC's: stratified cross-validation to compare, ranking by balanced accuracy \citep{brodersen2010balanced}, and nested cross-validation or an untouched stratified 20\% test set for the final score; the most-frequent-class rule is shown for reference.

\section{The desktop application}
\label{sec:desktop}

\ER{} Desktop is distributed as a Windows setup program that installs the application together with .NET and an isolated Python runtime, so neither Python nor administrator rights are needed. A sidebar lists the four tasks. After a file is read, ``What we noticed'' explains in plain sentences what the analysis will do with it (columns left out and why, empty cells, repeated rows, rare classes, how the final score will be obtained, the length and season of a series). A ``Check and run'' step summarises the analysis and enables Run only when nothing is missing. Advanced options, closed by default, restore \EC's expert choices. Problems such as a damaged file or a file open in another program are explained in one sentence. Every results folder contains the report, tables, predictions or forecasts, figures, the trained model, \texttt{settings.json} (what was asked and what was used), \texttt{versions.txt} (every package version) and a log.

The desktop and the cores are separated by a versioned protocol (\cref{fig:architecture}): each job runs in its own Python process, writes JSON-lines events, and can be cancelled by terminating the process tree. Labels, measures, result tiles and evaluation options come from each module's description of itself, so no screen is specific to one task.

\section{Illustrative examples}
\label{sec:examples}

\begin{figure}[t]
\centering
\begin{minipage}[b]{0.31\linewidth}\centering
\includegraphics[width=\linewidth]{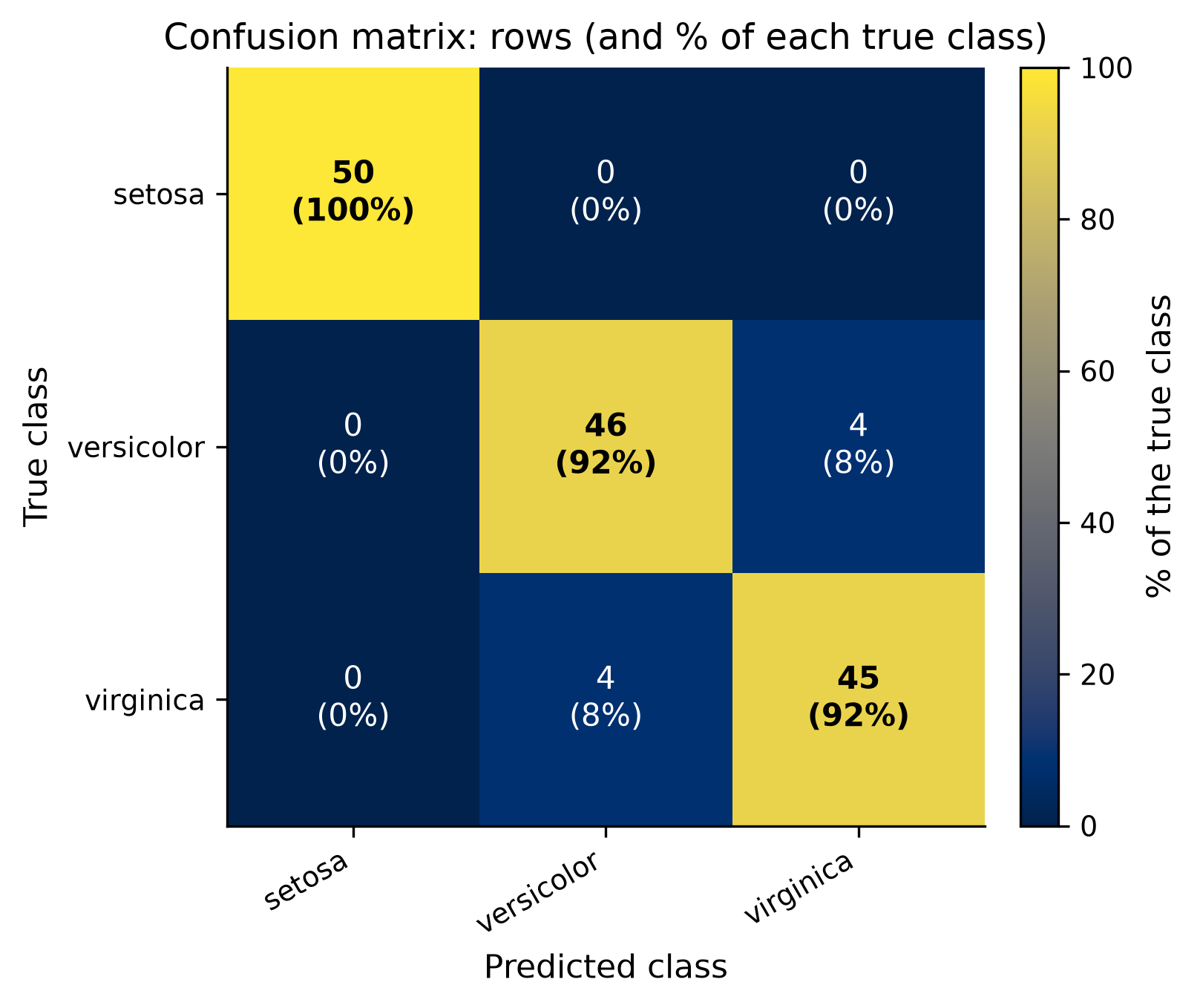}\\[-1mm]{\small (a) Classification}
\end{minipage}\hfill
\begin{minipage}[b]{0.29\linewidth}\centering
\includegraphics[width=\linewidth]{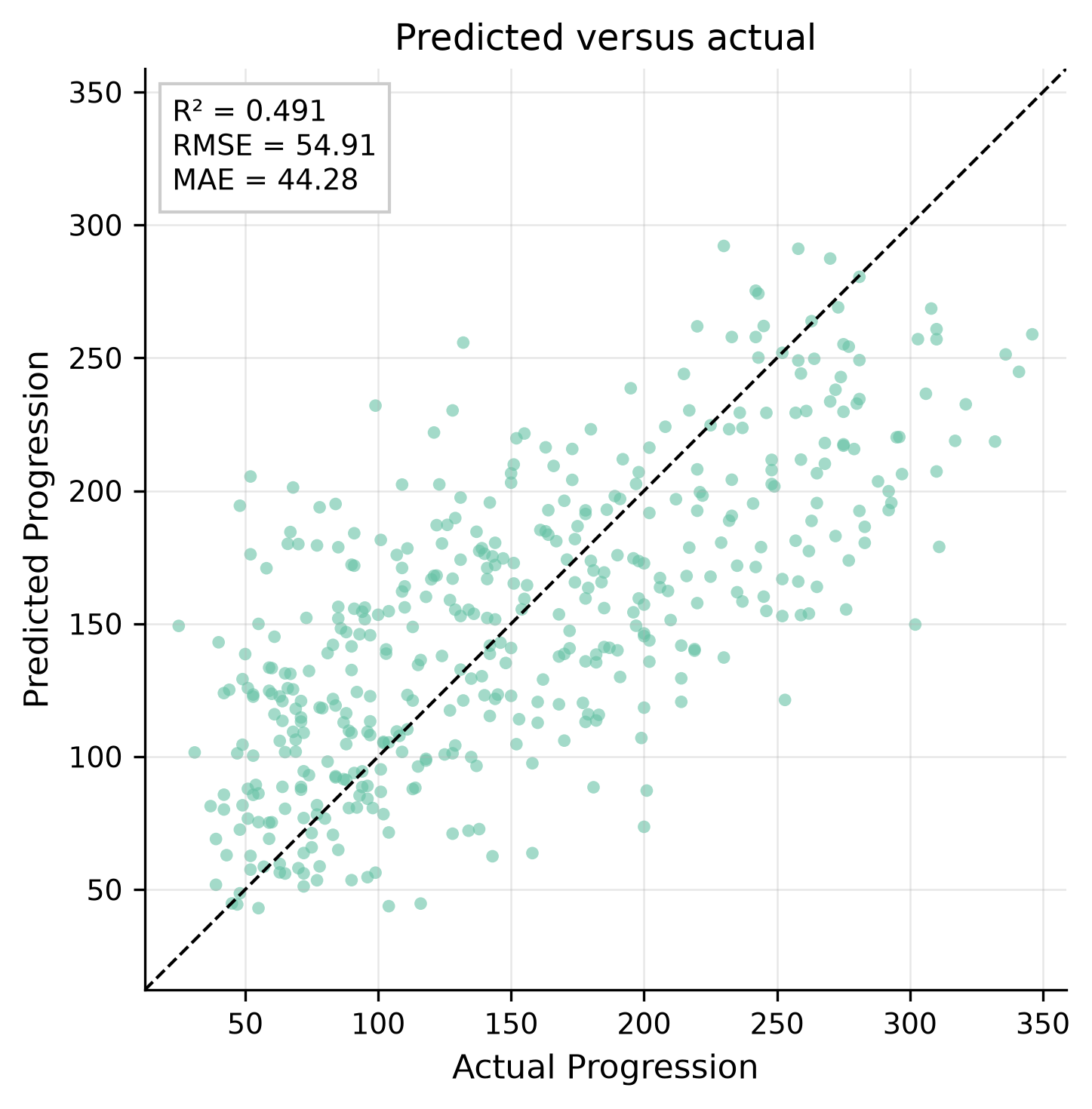}\\[-1mm]{\small (b) Regression}
\end{minipage}\hfill
\begin{minipage}[b]{0.31\linewidth}\centering
\includegraphics[width=\linewidth]{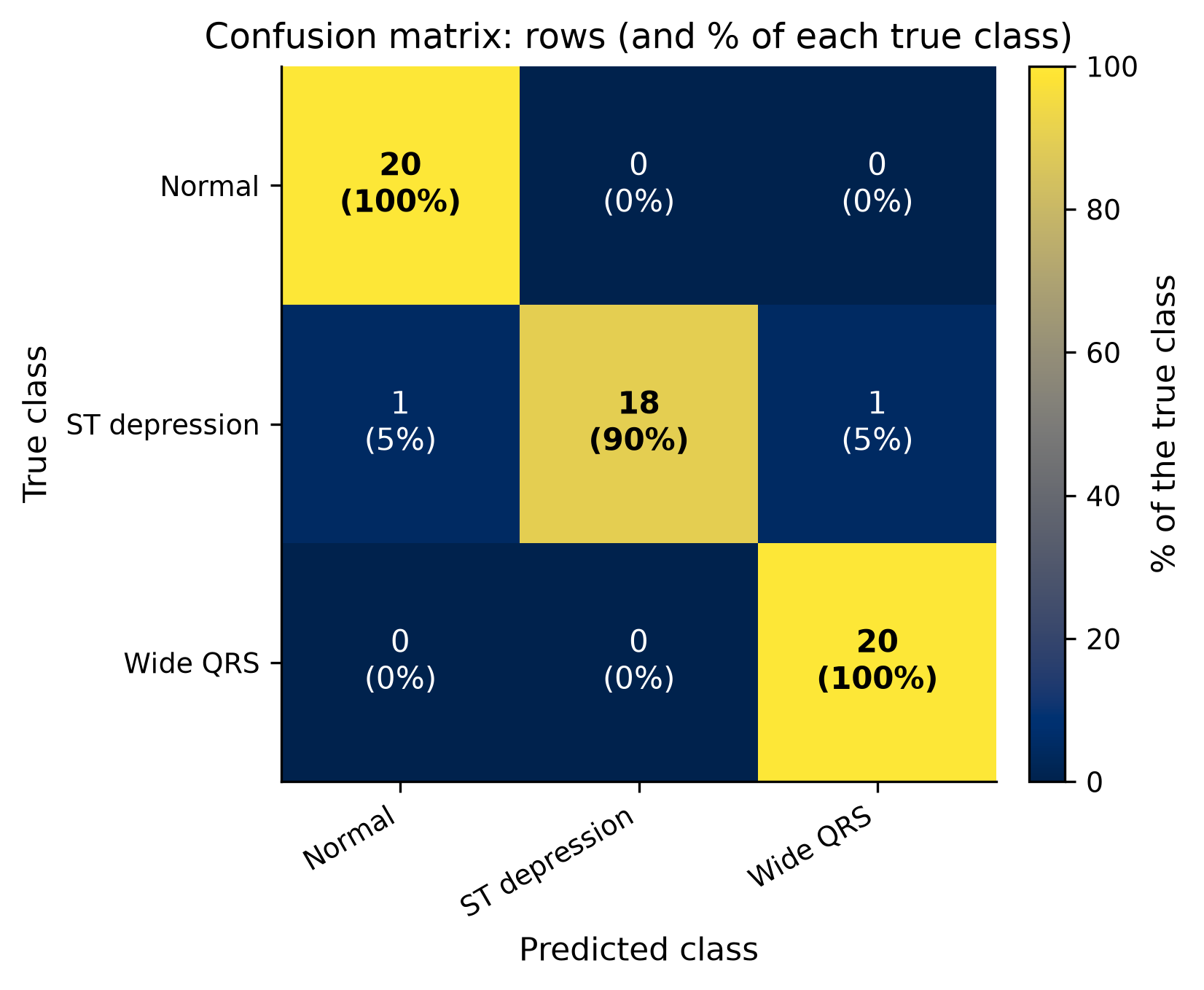}\\[-1mm]{\small (c) Signals}
\end{minipage}\\[3mm]
\begin{minipage}[b]{0.85\linewidth}\centering
\includegraphics[width=\linewidth]{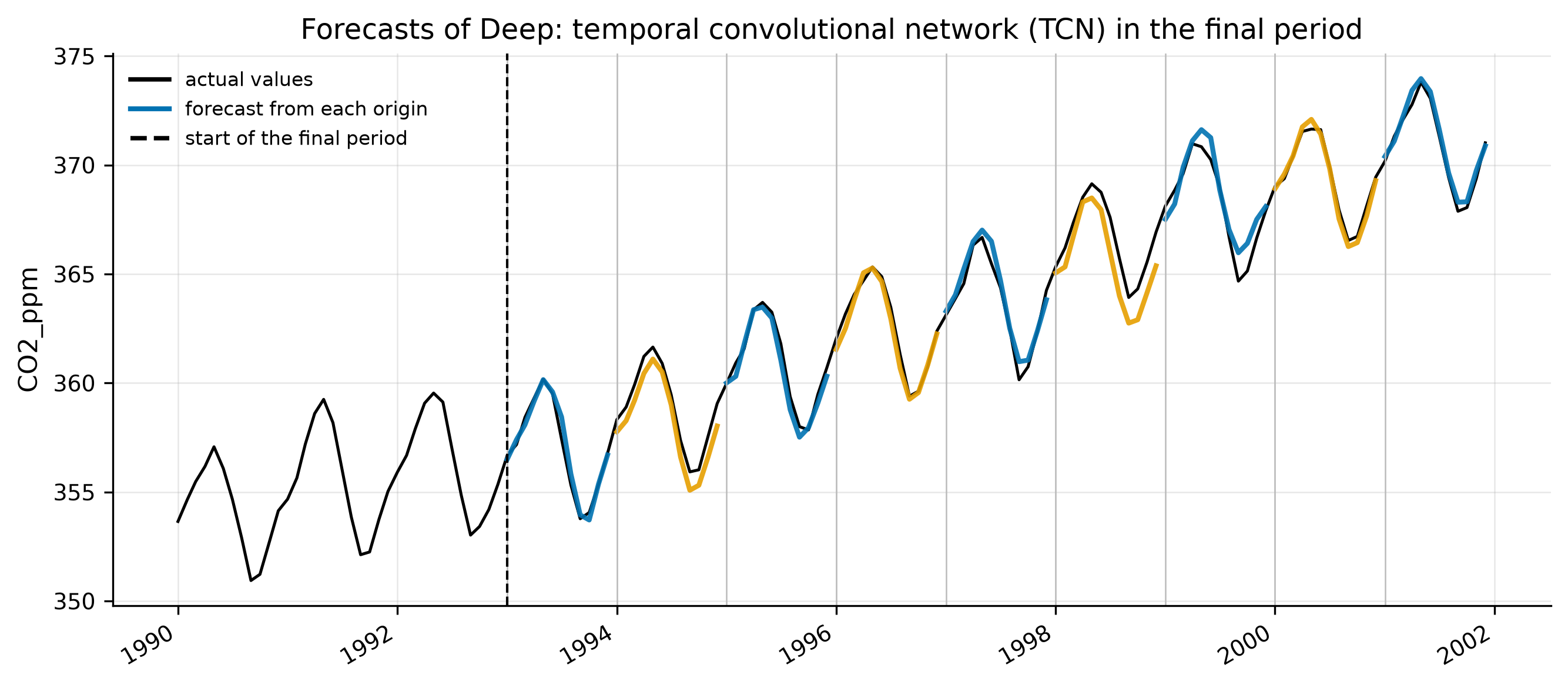}\\[-1mm]{\small (d) Forecasting}
\end{minipage}
\caption{One figure from each example report, as drawn by the software. (a) Iris: confusion matrix of the nested cross-validation predictions. (b) Diabetes: predicted against actual progression on held-out rows. (c) Synthetic heartbeats: confusion matrix on the 60 untouched recordings. (d) Mauna Loa CO$_2$: 12-month forecasts of the selected model from each origin in the final period, which played no part in choosing the model; consecutive forecasts alternate between blue and orange, and thin grey lines mark the origins.}
\label{fig:examples}
\end{figure}

Each task in \ER{} Desktop has an example button that loads a dataset shipped with the application. \Cref{tab:examples} summarises one run of each, made with the default validation choices and, except for classification, every available model including the optional deep networks; \cref{fig:examples} shows one figure from each report. The runs were made with \ER{} Desktop 0.1.2; the classification run was made with an earlier build using \EC{} 0.8.1, and repeating it with \EC{} 0.8.2 gave identical results. Each run took a few clicks: choose the task and the example, accept the suggested target, select the models and press Run.

\begin{table}[t]
\centering\footnotesize
\caption{One run of each example. \emph{Comparison}: score of the selected model in the comparison used to choose it; \emph{Final}: the score \ER{} reports, from data not used to choose the model; \emph{Reference}: the most-frequent-class rule (classification and signals), the training mean (regression) and the seasonal naive rule in the final period (forecasting). Classification and signals: balanced accuracy (\%); regression: $R^2$; forecasting: MASE (lower is better).}
\label{tab:examples}
\setlength{\tabcolsep}{4pt}
\begin{tabular}{@{}lllrrr@{}}
\toprule
Task and data & Models & Selected model (final evaluation) & Comparison & Final & Reference \\
\midrule
Classification: Iris, 149 rows & 3 & Logistic regression (nested CV) & 96.6 & 94.6 & 33.3 \\
Regression: diabetes, 442 rows & 10 & Ridge regression (nested CV) & 0.492 & 0.491 & $-0.007$ \\
Forecasting: CO$_2$, 526 months & 12 + 2 rules & TCN (last 108 months) & 0.365 & 0.368 & 1.283 \\
Signals: heartbeats, 300 recordings & 6 & ROCKET (untouched 20\%) & 98.3 & 96.7 & 33.3 \\
\bottomrule
\end{tabular}
\end{table}

\paragraph{Classification.} On Iris \citep{fisher1936iris} the tool removed one repeated row and compared a decision tree, a random forest and logistic regression by stratified 5-fold cross-validation. Logistic regression had the highest balanced accuracy, 96.6\%; nested cross-validation, in which logistic regression was selected in three of the five outer folds, gave the reported 94.6\%, 2.0 points less. The report adds that the lead over the decision tree (1.3 points) was smaller than the variation between folds (3.1 points), so the choice between the top models is not clear-cut.

\paragraph{Regression.} On the diabetes data \citep{efron2004lars} all ten models were compared. Ridge and linear regression were best and equal to three decimals; the reported nested score was $R^2 = 0.491$, with an RMSE 29\% smaller than always predicting the training mean (54.9 against 77.3) and an MHSP of 74.5\% against 65.3\%. The decision tree ($R^2 = -0.20$) did worse than the mean, which the reference row makes visible.

\paragraph{Forecasting.} The monthly Mauna Loa CO$_2$ series distributed with statsmodels \citep{seabold2010statsmodels} has 526 months, five of which were missing and filled with the last known value. The tool chose a horizon of 12 months and a season of 12, compared twelve models and the two reference rules at rolling origins in the first 418 months, and kept the last 108 months (nine forecasts of 12 months) for the final evaluation. A temporal convolutional network \citep{bai2018tcn} had the lowest comparison MASE (0.365), and its final-period MASE was 0.368, against 1.283 for the seasonal naive rule; the 80\% band contained 75\% of the actual values. The report also states that its lead over the GRU (0.367) was smaller than the variation between forecast origins (0.251): a ridge model on past values (0.375) would have served almost equally well.

\paragraph{Signals.} The synthetic heartbeats (300 recordings of 140 values in three shape classes, generated for trying the tool and not real patient data) were classified with six models. Because deep networks were included, 60 recordings (20\%) were set aside before the comparison. ROCKET won the comparison with 98.3\% balanced accuracy and scored 96.7\% on the untouched recordings (two errors, \cref{fig:examples}c); InceptionTime was close (97.1\%), while the LSTM (42.1\%) was barely above the 33.3\% of the most-frequent-class rule.

\paragraph{Reports.} Each run also wrote a report whose Methods paragraph can be adapted for a paper. For the regression example it begins: ``Regression was carried out with EasyResearch (version 0.1.2), which uses the data preparation of EasyClassifier and is built on scikit-learn, NumPy and pandas. \ldots{} All learned preparation steps were fitted on the training rows of each split only, so no information from test rows reached the models'', and goes on to name every model, the evaluation design and the measures, each with its reference.

\section{Evaluation}
\label{sec:evaluation}

\subsection{Correctness against reference implementations}
Each core was first checked against independent code on public data (\cref{tab:agreement}): the cross-validated $R^2$ of linear regression against a plain scikit-learn pipeline on the same folds, the final-period error of the seasonal naive rule recomputed from the raw series, and \ER's ROCKET against aeon's implementation on the official UCR train/test splits.

\begin{table}[t]
\centering\small
\caption{Agreement with reference implementations (largest absolute difference over the datasets).}
\label{tab:agreement}
\begin{tabular}{@{}llr@{}}
\toprule
Task & Reference & Largest difference \\
\midrule
Regression (5 datasets) & linear-regression $R^2$, scikit-learn pipeline & 0.0000 \\
Forecasting (6 series) & seasonal-naive MAE, recomputed from the series & 0.0000 \\
Signals (4 UCR datasets) & ROCKET accuracy, aeon \texttt{RocketClassifier} & 0.0171 \\
\bottomrule
\end{tabular}
\end{table}

ROCKET draws its kernels at random, so two implementations are not expected to agree exactly; the largest difference was on ArrowHead (80.6\% against 82.3\% accuracy), and the other three datasets agreed exactly.

\subsection{Are the reported scores honest?}

\paragraph{Protocol.} We repeated the design of \EC's honesty benchmark \citep{hassanat2026easyclassifierpaper}. For regression and signals, each dataset was split in half at random (\RegSeeds{} splits for regression, \SigSeeds{} for signals); the first half was analysed with all available standard models (regression) or the default set (signals), and the second half, never touched, gave the true performance of the model each estimate describes, refitted on the whole first half. For forecasting, the first 70\% of each series was the development part and the last 30\% the untouched future, in which the selected model was evaluated at rolling origins one horizon apart, retrained before each origin. The estimates compared with the truth were:
\begin{itemize}
  \item \emph{Common practice}: for regression, preprocessing fitted on all development rows before 5-fold cross-validation and the best model's score reported; for forecasting, the machine-learning models scored by shuffled 5-fold cross-validation over the windows of the development part and the best one reported;
  \item \emph{Comparison best}: the best score in \ER's own leakage-free comparison (selection bias only);
  \item \emph{Final score}: the score \ER{} reports.
\end{itemize}
Controls were a random target (regression; 300 rows, 10 noise predictors), random labels (signals; 200 random-walk recordings) and \FcRWn{} random walks (forecasting), where nothing can be predicted. The benchmark code and all results are in the repository.

\paragraph{Datasets.} Regression: diabetes progression \citep{efron2004lars}, occupational prestige \citep{fox2011car}, New York air quality \citep{chambers1983graphical}, Friedman~\#1 \citep{friedman1991mars}, professors' salaries \citep{fox2011car}, car highway mileage, interlocking directorates \citep{fox2011car}, computer performance, and wages in the Survey of Labour and Income Dynamics \citep{fox2011car}; the last has more than 2{,}000 rows in the development half, so its final score came from an untouched 20\% test set. Forecasting: \FcNData{} monthly, quarterly, yearly and undated series distributed with statsmodels and R, including Mauna Loa CO$_2$, airline passengers \citep{box1976}, UK gas consumption \citep{durbin2001} and Nile flow \citep{cobb1978nile}. Signals: GunPoint, ItalyPowerDemand, ArrowHead and OSULeaf from the UCR archive \citep{dau2019ucr}.

\begin{table}[t]
\centering\small
\caption{Regression: $R^2$ ($\times 100$) estimated on the first half and measured on the untouched second half (mean of 10 random splits). The truth is that of \ER's selected model; the common-practice estimate is compared with its own winner's truth when its bias is computed.}
\label{tab:regression}
\resizebox{\linewidth}{!}{%
\begin{tabular}{@{}lrrrrr@{}}
\toprule
Dataset & Rows & Common practice & Comparison best & \ER{} final & Truth \\
\midrule
Diabetes progression & 221 & 49.4 & 49.2 & 45.9 & 46.6 \\
Occupational prestige (Canada) & 51 & 81.9 & 81.5 & 76.8 & 73.6 \\
New York air quality (ozone) & 58 & 63.9 & 64.9 & 60.4 & 59.5 \\
Friedman \#1 (synthetic, noise sd 1) & 250 & 81.0 & 81.2 & 81.0 & 82.4 \\
Professors' salaries (US) & 196 & 43.0 & 43.6 & 41.0 & 39.8 \\
Car highway mileage (mpg) & 109 & 87.0 & 86.4 & 84.0 & 83.5 \\
Interlocking directorates (Canada) & 124 & 53.4 & 54.9 & 46.7 & 56.4 \\
Computer performance (cpus) & 104 & 86.6 & 87.6 & 80.7 & 83.5 \\
Survey of labour and income (wages) & 2055 & 34.5 & 33.6 & 35.1 & 35.2 \\
\midrule
Random target (control) & 150 & $-$6.1 & $-$4.3 & $-$8.9 & $-$15.7 \\
\bottomrule
\end{tabular}}
\end{table}

\begin{figure}[t]
\centering
\includegraphics[width=\linewidth]{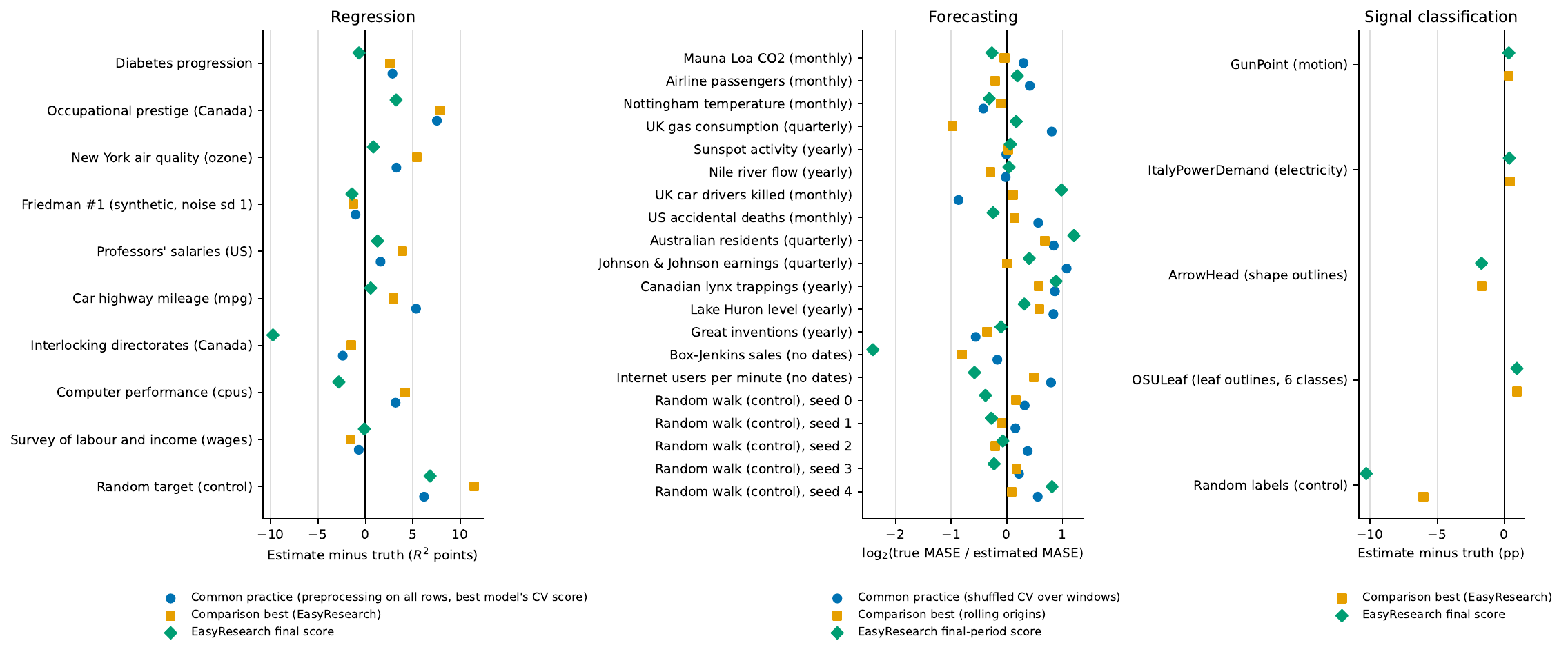}
\caption{Bias of each estimate against the untouched data, per dataset (regression and signals: mean over random splits; forecasting: one development--future split per series). Positive values are optimistic: the estimate promised more than was achieved. Forecasting is shown on the log scale of the ratio of achieved to estimated MASE, so $+1$ means the achieved error was twice the estimate.}
\label{fig:honesty}
\end{figure}

\paragraph{Regression.} \Cref{tab:regression,fig:honesty} show the results. Over the \RegRuns{} runs on real data, the common practice overstated $R^2$ by \RegBiasNaive{} points on average and the best comparison score by \RegBiasComp, while \ER's final score was \RegBiasFinal{} points from the truth, slightly conservative because nested cross-validation trains on fewer rows than the final model (one-sided Wilcoxon signed-rank test that the common practice is more optimistic than the final score: $p = \RegP$; averaging the splits within each dataset first: $p = \RegPds$). On the small halves every estimate was noisy: the mean absolute errors were \RegMAENaive, \RegMAEComp{} and \RegMAEFinal{} points, so the final score removes the optimism but cannot remove the uncertainty of a small sample. Fitting preprocessing on all rows added little on its own (common practice and comparison best were close), as in the classification benchmark; the optimism came from reporting the winner's own score. On the random target, where nothing can be predicted, the best comparison score was $R^2 = \RegCtrlCompVal$, the final score \RegCtrlFinalVal{} and the truth \RegCtrlTruthVal; both reports were negative, which the report states as ``no better than always using the average''.

\paragraph{Screening on all rows.} With 100 rows, 500 noise predictors and a noise outcome, selecting the 10 predictors most correlated with the outcome on all rows and then cross-validating a linear model reported $R^2 = \ScrOut$ on average over \RegSeeds{} draws; repeating the selection inside each fold gave \ScrIn, and the true value on new data was \ScrExt. \ER{} offers no feature selection, partly for this reason.

\begin{table}[t]
\centering\footnotesize
\caption{Forecasting: MASE estimated on the development part (first 70\%) and achieved in the untouched future (last 30\%) by the model each procedure selected; lower is better. $H$ is the horizon. The last column is the naive or seasonal naive rule in the future.}
\label{tab:forecasting}
\resizebox{\linewidth}{!}{%
\begin{tabular}{@{}lrrrrrr@{}}
\toprule
 & & \multicolumn{2}{c}{Shuffled CV} & \multicolumn{2}{c}{\ER} & Rule \\
\cmidrule(lr){3-4}\cmidrule(lr){5-6}
Series & $H$ & estimate (model) & truth & final (model) & truth & truth \\
\midrule
Mauna Loa CO2 (monthly) & 12 & 0.361 (KNN) & 0.444 & 0.447 (Ridge) & 0.371 & 1.204 \\
Airline passengers (monthly) & 10 & 0.396 (Ridge) & 0.527 & 0.462 (Ridge) & 0.527 & 1.275 \\
Nottingham temperature (monthly) & 12 & 0.792 (Ridge) & 0.591 & 0.757 (ETS) & 0.609 & 0.877 \\
UK gas consumption (quarterly) & 4 & 0.927 (Ridge) & 1.622 & 1.443 (Ridge) & 1.622 & 1.740 \\
Sunspot activity (yearly) & 5 & 1.075 (RF) & 1.069 & 1.022 (RF) & 1.069 & 3.432 \\
Nile river flow (yearly) & 5 & 1.008 (RF) & 0.993 & 0.889 (Ridge) & 0.915 & 0.861 \\
UK car drivers killed (monthly) & 12 & 1.100 (RF) & 0.603 & 0.458 (ETS) & 0.904 & 1.216 \\
US accidental deaths (monthly) & 5 & 1.169 (RF) & 1.729 & 0.607 (seas.\ naive) & 0.512 & 0.512 \\
Australian residents (quarterly) & 4 & 0.072 (Ridge) & 0.129 & 0.052 (ETS) & 0.121 & 1.211 \\
Johnson \& Johnson earnings (quarterly) & 4 & 0.587 (Ridge) & 1.237 & 0.994 (ETS) & 1.314 & 2.627 \\
Canadian lynx trappings (yearly) & 8 & 0.891 (RF) & 1.623 & 0.879 (RF) & 1.623 & 2.759 \\
Lake Huron level (yearly) & 6 & 1.451 (RF) & 2.593 & 2.086 (RF) & 2.593 & 2.243 \\
Great inventions (yearly) & 7 & 0.999 (RF) & 0.678 & 0.708 (Theta) & 0.659 & 0.912 \\
Box-Jenkins sales (no dates) & 10 & 3.428 (KNN) & 3.048 & 7.350 (ETS) & 1.387 & 1.851 \\
Internet users per minute (no dates) & 7 & 2.969 (RF) & 5.149 & 5.057 (ETS) & 3.381 & 4.819 \\
\midrule
Random walk (control), seed 0 & 10 & 1.879 (KNN) & 2.349 & 2.378 (naive) & 1.823 & 1.823 \\
Random walk (control), seed 1 & 10 & 2.280 (KNN) & 2.536 & 2.574 (naive) & 2.125 & 2.125 \\
Random walk (control), seed 2 & 10 & 1.983 (RF) & 2.572 & 2.654 (naive) & 2.527 & 2.527 \\
Random walk (control), seed 3 & 10 & 2.306 (RF) & 2.686 & 2.742 (naive) & 2.341 & 2.341 \\
Random walk (control), seed 4 & 10 & 2.150 (RF) & 3.162 & 1.543 (ETS) & 2.711 & 2.650 \\
\bottomrule
\end{tabular}}
\end{table}

\paragraph{Forecasting.} \Cref{tab:forecasting} shows the results as MASE. Because relative errors are asymmetric, we summarise the ratio of estimated to achieved MASE on the log scale. Shuffled cross-validation promised a smaller error than its selected model achieved in the future on \FcShufBelow{} of \FcNData{} series; on average (geometric mean) the promised error was \FcShufGeo\% too small (median \FcShufMedian\%). \ER's final-period score was unbiased on average (geometric mean within \FcFinalGeo\% of the truth) but, measured on a short final period, it was as variable as the shuffled estimate: a typical estimate was off by a factor of \FcFinalTyp{} (shuffled \FcShufTyp, comparison best \FcCompTyp), and the difference in bias was not significant over \FcNData{} series ($p = \FcP$). The choice of model mattered more than the estimate: the models \ER{} selected by rolling-origin comparison had a lower future error than those selected by shuffled cross-validation on \FcToolBetter{} series, the same model on \FcSame{} and a higher error on \FcToolWorse{} (\FcTruthGain\% lower on average), and they beat the naive or seasonal naive rule in the future on \FcBeatsRule{} series against \FcBeatsRuleShuf. On the random walks, shuffled cross-validation selected a machine-learning model in \FcRWShufML{} of \FcRWn{} series and promised an error \FcRWShufGain\% below what it achieved; \FcRWShufWorse{} of these models were worse in the future than repeating the last value. \ER{} selected the naive rule itself on \FcRWToolNaive{} of the \FcRWn{} random walks.

\begin{table}[t]
\centering\small
\caption{Signal classification: balanced accuracy (\%) estimated on the first half and measured on the untouched second half (mean of 3 random stratified splits), default model set.}
\label{tab:signals}
\begin{tabular}{@{}lrrrr@{}}
\toprule
Dataset & Recordings & Comparison best & \ER{} final & Truth \\
\midrule
GunPoint (motion) & 100 & 100.0 & 100.0 & 99.7 \\
ItalyPowerDemand (electricity) & 548 & 97.4 & 97.4 & 97.0 \\
ArrowHead (shape outlines) & 105 & 93.4 & 93.4 & 95.1 \\
OSULeaf (leaf outlines, 6 classes) & 221 & 94.0 & 94.0 & 93.0 \\
\midrule
Random labels (control) & 100 & 46.9 & 42.7 & 52.9 \\
\bottomrule
\end{tabular}
\end{table}

\paragraph{Signals.} \Cref{tab:signals} shows the results. ROCKET won the comparison in all \SigRuns{} runs on real data, so there was no selection to correct: the best comparison score and the final score were both unbiased on average (\SigBiasComp{} and \SigBiasFinal{} points; mean absolute errors \SigMAEComp{} and \SigMAEFinal). On random labels the comparison best reported \SigCtrlComp\% balanced accuracy, the final score \SigCtrlFinal\% and the truth on the untouched half was \SigCtrlTruth\%; all three are chance level within the noise of 100 recordings. The final score therefore costs computing time without changing the result when one method clearly dominates, and protects the user when it does not; the software cannot know in advance which case applies.

\section{Impact, limitations and future work}
\label{sec:impact}

\paragraph{Impact.} \ER{} brings honest regression, forecasting and signal classification to researchers who do not program, in the same application as \EC's classification, with plain-language explanations and reports that a reviewer can check. The benchmark shows where the protection matters. Shuffled evaluation of forecasts promised errors about a fifth smaller than were achieved and, on series where nothing can be forecast, preferred machine-learning models that lost to the simplest rule; in regression, reporting the winner's comparison score was optimistic. Where one method clearly dominates, as ROCKET did on the signal datasets, the honest final score changes little. In every task, estimates from small samples remained noisy, which a tool can report but not remove.

\paragraph{Limitations.} Signal classification handles one channel per recording; several channels and long recordings cut into windows are planned. Forecasting handles one series without explanatory variables. Several rows per subject (repeated measurements) are not yet grouped in cross-validation, so users with such data must aggregate first. The desktop application runs on Windows only. Deep models are slow on a normal computer and are therefore optional. The benchmark used modest datasets, a single development--future split per series for forecasting and three splits per dataset for signals; it shows the direction and order of the biases, not their exact size elsewhere.

\paragraph{Future work.} Computer vision will be a separate application, \EV, built on the same rules, with subject-wise splitting, augmentation confined to training images and a mandatory check for copied images. For \ER, grouped cross-validation, multichannel signals and a server version are planned.

\section{Conclusions}
\ER{} extends the rules of \EC{} to regression, forecasting and signal classification and brings all four tasks into one application for researchers who do not program. Its reported scores were free of the optimism of the common practice in regression and forecasting, unbiased where there was nothing to correct, and on data where nothing can be predicted it said so.

\section*{Acknowledgements}
The authors thank the developers of scikit-learn, statsmodels, PyTorch, aeon, pandas, NumPy and matplotlib, and the authors of the public datasets used in the evaluation. Parts of the software, the benchmark scripts and this manuscript were prepared with the help of Claude (Anthropic), an AI assistant, which was used for coding support, drafting and editing. The authors designed the software and the study, checked all code, results and text, and take full responsibility for the content of this paper.

\section*{Software and data availability}
{\sloppy
\begin{itemize}
  \item Source code of \ER{} and \ER{} Desktop: \url{https://github.com/ahmadhassant/easyresearch}; archived at Zenodo, \TBDDOI.
  \item \EC: \url{https://github.com/ahmadhassant/easyclassifier}; Zenodo, \href{https://doi.org/10.5281/zenodo.23122901}{doi:10.5281/zenodo.23122901} (all versions).
  \item Benchmarks: \nolinkurl{benchmarks/easyresearch_benchmarks.py} and \nolinkurl{benchmarks/easyresearch_honesty.py} in the same repository, with all results; every dataset is installed with pip.
\end{itemize}}

\bibliographystyle{plainnat}
\bibliography{references}

\begin{thebibliography}{36}
\providecommand{\natexlab}[1]{#1}
\providecommand{\url}[1]{\texttt{#1}}
\expandafter\ifx\csname urlstyle\endcsname\relax
  \providecommand{\doi}[1]{doi: #1}\else
  \providecommand{\doi}{doi: \begingroup \urlstyle{rm}\Url}\fi

\bibitem[Alaydaa et~al.(2026)Alaydaa, Alotibi, Tarawneh, and
  Hassanat]{hassanat2014distance2}
Mohammad~Saad Alaydaa, Gaseb~N. Alotibi, Ahmad~S. Tarawneh, and Ahmad~B.
  Hassanat.
\newblock A sign-symmetric reformulation of the {H}assanat distance for data
  with negative feature values.
\newblock \emph{Symmetry}, 18\penalty0 (7), 2026.
\newblock ISSN 2073-8994.
\newblock \doi{10.3390/sym18071225}.
\newblock URL \url{https://www.mdpi.com/2073-8994/18/7/1225}.

\bibitem[Assimakopoulos and Nikolopoulos(2000)]{assimakopoulos2000theta}
Vassilis Assimakopoulos and Konstantinos Nikolopoulos.
\newblock The theta model: a decomposition approach to forecasting.
\newblock \emph{International Journal of Forecasting}, 16\penalty0
  (4):\penalty0 521--530, 2000.

\bibitem[Bai et~al.(2018)Bai, Kolter, and Koltun]{bai2018tcn}
Shaojie Bai, J.~Zico Kolter, and Vladlen Koltun.
\newblock An empirical evaluation of generic convolutional and recurrent
  networks for sequence modeling, 2018.
\newblock arXiv:1803.01271.

\bibitem[Bergmeir and Ben{\'\i}tez(2012)]{bergmeir2012cv}
Christoph Bergmeir and Jos{\'e}~M. Ben{\'\i}tez.
\newblock On the use of cross-validation for time series predictor evaluation.
\newblock \emph{Information Sciences}, 191:\penalty0 192--213, 2012.
\newblock \doi{10.1016/j.ins.2011.12.028}.

\bibitem[Bergmeir et~al.(2018)Bergmeir, Hyndman, and Koo]{bergmeir2018note}
Christoph Bergmeir, Rob~J. Hyndman, and Bonsoo Koo.
\newblock A note on the validity of cross-validation for evaluating
  autoregressive time series prediction.
\newblock \emph{Computational Statistics \& Data Analysis}, 120:\penalty0
  70--83, 2018.
\newblock \doi{10.1016/j.csda.2017.11.003}.

\bibitem[Bontempi et~al.(2013)Bontempi, Ben~Taieb, and
  Le~Borgne]{bontempi2013strategies}
Gianluca Bontempi, Souhaib Ben~Taieb, and Yann-A{\"e}l Le~Borgne.
\newblock Machine learning strategies for time series forecasting.
\newblock In \emph{Business Intelligence (eBISS 2012)}, volume 138 of
  \emph{Lecture Notes in Business Information Processing}, pages 62--77.
  Springer, 2013.
\newblock \doi{10.1007/978-3-642-36318-4_3}.

\bibitem[Box and Jenkins(1976)]{box1976}
George E.~P. Box and Gwilym~M. Jenkins.
\newblock \emph{Time Series Analysis: Forecasting and Control}.
\newblock Holden-Day, revised edition, 1976.

\bibitem[Breiman(2001)]{breiman2001rf}
Leo Breiman.
\newblock Random forests.
\newblock \emph{Machine Learning}, 45\penalty0 (1):\penalty0 5--32, 2001.
\newblock \doi{10.1023/A:1010933404324}.

\bibitem[Brodersen et~al.(2010)Brodersen, Ong, Stephan, and
  Buhmann]{brodersen2010balanced}
Kay~Henning Brodersen, Cheng~Soon Ong, Klaas~Enno Stephan, and Joachim~M.
  Buhmann.
\newblock The balanced accuracy and its posterior distribution.
\newblock In \emph{Proceedings of the 20th International Conference on Pattern
  Recognition (ICPR)}, pages 3121--3124, 2010.
\newblock \doi{10.1109/ICPR.2010.764}.

\bibitem[Cawley and Talbot(2010)]{cawley2010overfitting}
Gavin~C. Cawley and Nicola L.~C. Talbot.
\newblock On over-fitting in model selection and subsequent selection bias in
  performance evaluation.
\newblock \emph{Journal of Machine Learning Research}, 11:\penalty0 2079--2107,
  2010.

\bibitem[Chambers et~al.(1983)Chambers, Cleveland, Kleiner, and
  Tukey]{chambers1983graphical}
John~M. Chambers, William~S. Cleveland, Beat Kleiner, and Paul~A. Tukey.
\newblock \emph{Graphical Methods for Data Analysis}.
\newblock Wadsworth, 1983.

\bibitem[Cobb(1978)]{cobb1978nile}
George~W. Cobb.
\newblock The problem of the {N}ile: conditional solution to a changepoint
  problem.
\newblock \emph{Biometrika}, 65\penalty0 (2):\penalty0 243--251, 1978.

\bibitem[Dau et~al.(2019)Dau, Bagnall, Kamgar, Yeh, Zhu, Gharghabi,
  Ratanamahatana, and Keogh]{dau2019ucr}
Hoang~Anh Dau, Anthony Bagnall, Kaveh Kamgar, Chin-Chia~Michael Yeh, Yan Zhu,
  Shaghayegh Gharghabi, Chotirat~Ann Ratanamahatana, and Eamonn Keogh.
\newblock The {UCR} time series archive.
\newblock \emph{IEEE/CAA Journal of Automatica Sinica}, 6\penalty0
  (6):\penalty0 1293--1305, 2019.
\newblock \doi{10.1109/JAS.2019.1911747}.

\bibitem[Dempster et~al.(2020)Dempster, Petitjean, and
  Webb]{dempster2020rocket}
Angus Dempster, Fran{\c{c}}ois Petitjean, and Geoffrey~I. Webb.
\newblock {ROCKET}: exceptionally fast and accurate time series classification
  using random convolutional kernels.
\newblock \emph{Data Mining and Knowledge Discovery}, 34\penalty0 (5):\penalty0
  1454--1495, 2020.

\bibitem[Durbin and Koopman(2001)]{durbin2001}
James Durbin and Siem~Jan Koopman.
\newblock \emph{Time Series Analysis by State Space Methods}.
\newblock Oxford University Press, 2001.

\bibitem[Efron et~al.(2004)Efron, Hastie, Johnstone, and
  Tibshirani]{efron2004lars}
Bradley Efron, Trevor Hastie, Iain Johnstone, and Robert Tibshirani.
\newblock Least angle regression.
\newblock \emph{The Annals of Statistics}, 32\penalty0 (2):\penalty0 407--499,
  2004.

\bibitem[Fisher(1936)]{fisher1936iris}
Ronald~A. Fisher.
\newblock The use of multiple measurements in taxonomic problems.
\newblock \emph{Annals of Eugenics}, 7\penalty0 (2):\penalty0 179--188, 1936.
\newblock \doi{10.1111/j.1469-1809.1936.tb02137.x}.

\bibitem[Fox and Weisberg(2011)]{fox2011car}
John Fox and Sanford Weisberg.
\newblock \emph{An {R} Companion to Applied Regression}.
\newblock Sage, 2nd edition, 2011.

\bibitem[Friedman(1991)]{friedman1991mars}
Jerome~H. Friedman.
\newblock Multivariate adaptive regression splines.
\newblock \emph{The Annals of Statistics}, 19\penalty0 (1):\penalty0 1--67,
  1991.

\bibitem[Fulcher and Jones(2014)]{fulcher2014features}
Ben~D. Fulcher and Nick~S. Jones.
\newblock Highly comparative feature-based time-series classification.
\newblock \emph{IEEE Transactions on Knowledge and Data Engineering},
  26\penalty0 (12):\penalty0 3026--3037, 2014.
\newblock \doi{10.1109/TKDE.2014.2316504}.

\bibitem[Hassanat et~al.(2022)Hassanat, Alkafaween, Tarawneh, and
  Elmougy]{hassanat2014distance}
Ahmad Hassanat, Esra'a Alkafaween, Ahmad~S. Tarawneh, and Samir Elmougy.
\newblock Applications review of {H}assanat distance metric.
\newblock In \emph{2022 International Conference on Emerging Trends in
  Computing and Engineering Applications (ETCEA)}, pages 1--6, 2022.
\newblock \doi{10.1109/ETCEA57049.2022.10009844}.

\bibitem[Hassanat(2014)]{hassanat2014arxiv}
Ahmad~B. Hassanat.
\newblock Dimensionality invariant similarity measure, 2014.
\newblock arXiv:1409.0923.

\bibitem[Hassanat et~al.(2024)Hassanat, Alqaralleh, Tarawneh, Almohammadi,
  Alamri, Alzahrani, Altarawneh, and Alhalaseh]{hassanat2024mhsp}
Ahmad~B. Hassanat, Mohammad~Khaled Alqaralleh, Ahmad~S. Tarawneh, Khalid
  Almohammadi, Maha Alamri, Abdulkareem Alzahrani, Ghada~A. Altarawneh, and
  Rania Alhalaseh.
\newblock A novel outlier-robust accuracy measure for machine learning
  regression using a non-convex distance metric.
\newblock \emph{Mathematics}, 12\penalty0 (22):\penalty0 3623, 2024.
\newblock \doi{10.3390/math12223623}.

\bibitem[Hassanat and Altarawneh(2026)]{hassanat2026easyclassifierpaper}
Ahmad B.~A. Hassanat and Ghada~A. Altarawneh.
\newblock {EasyClassifier}: Honest, reproducible machine-learning
  classification for researchers who do not program, 2026.
\newblock arXiv:2610.04758.

\bibitem[Hastie et~al.(2009)Hastie, Tibshirani, and Friedman]{hastie2009esl}
Trevor Hastie, Robert Tibshirani, and Jerome Friedman.
\newblock \emph{The Elements of Statistical Learning: Data Mining, Inference,
  and Prediction}.
\newblock Springer, 2nd edition, 2009.
\newblock \doi{10.1007/978-0-387-84858-7}.

\bibitem[Hyndman and Athanasopoulos(2021)]{hyndman2021fpp3}
Rob~J. Hyndman and George Athanasopoulos.
\newblock \emph{Forecasting: Principles and Practice}.
\newblock OTexts, 3rd edition, 2021.
\newblock URL \url{https://otexts.com/fpp3/}.

\bibitem[Hyndman and Koehler(2006)]{hyndman2006mase}
Rob~J. Hyndman and Anne~B. Koehler.
\newblock Another look at measures of forecast accuracy.
\newblock \emph{International Journal of Forecasting}, 22\penalty0
  (4):\penalty0 679--688, 2006.

\bibitem[Hyndman et~al.(2002)Hyndman, Koehler, Snyder, and
  Grose]{hyndman2002ets}
Rob~J. Hyndman, Anne~B. Koehler, Ralph~D. Snyder, and Simone Grose.
\newblock A state space framework for automatic forecasting using exponential
  smoothing methods.
\newblock \emph{International Journal of Forecasting}, 18\penalty0
  (3):\penalty0 439--454, 2002.

\bibitem[Kapoor and Narayanan(2023)]{kapoor2023leakage}
Sayash Kapoor and Arvind Narayanan.
\newblock Leakage and the reproducibility crisis in machine-learning-based
  science.
\newblock \emph{Patterns}, 4\penalty0 (9):\penalty0 100804, 2023.
\newblock \doi{10.1016/j.patter.2023.100804}.

\bibitem[Lones(2021)]{lones2021pitfalls}
Michael~A. Lones.
\newblock How to avoid machine learning pitfalls: a guide for academic
  researchers, 2021.
\newblock arXiv:2108.02497.

\bibitem[Paszke et~al.(2019)]{paszke2019pytorch}
Adam Paszke et~al.
\newblock {PyTorch}: An imperative style, high-performance deep learning
  library.
\newblock In \emph{Advances in Neural Information Processing Systems 32}, pages
  8024--8035, 2019.

\bibitem[Pedregosa et~al.(2011)Pedregosa, Varoquaux, Gramfort, Michel, Thirion,
  Grisel, Blondel, Prettenhofer, Weiss, Dubourg, Vanderplas, Passos,
  Cournapeau, Brucher, Perrot, and Duchesnay]{pedregosa2011sklearn}
Fabian Pedregosa, Ga{\"e}l Varoquaux, Alexandre Gramfort, Vincent Michel,
  Bertrand Thirion, Olivier Grisel, Mathieu Blondel, Peter Prettenhofer, Ron
  Weiss, Vincent Dubourg, Jake Vanderplas, Alexandre Passos, David Cournapeau,
  Matthieu Brucher, Matthieu Perrot, and {\'E}douard Duchesnay.
\newblock Scikit-learn: Machine learning in {P}ython.
\newblock \emph{Journal of Machine Learning Research}, 12:\penalty0 2825--2830,
  2011.

\bibitem[Probst et~al.(2019)Probst, Boulesteix, and
  Bischl]{probst2019tunability}
Philipp Probst, Anne-Laure Boulesteix, and Bernd Bischl.
\newblock Tunability: Importance of hyperparameters of machine learning
  algorithms.
\newblock \emph{Journal of Machine Learning Research}, 20\penalty0
  (53):\penalty0 1--32, 2019.

\bibitem[Seabold and Perktold(2010)]{seabold2010statsmodels}
Skipper Seabold and Josef Perktold.
\newblock Statsmodels: Econometric and statistical modeling with {P}ython.
\newblock In \emph{Proceedings of the 9th Python in Science Conference}, pages
  92--96, 2010.

\bibitem[Tashman(2000)]{tashman2000}
Leonard~J. Tashman.
\newblock Out-of-sample tests of forecasting accuracy: an analysis and review.
\newblock \emph{International Journal of Forecasting}, 16\penalty0
  (4):\penalty0 437--450, 2000.

\bibitem[Varma and Simon(2006)]{varma2006bias}
Sudhir Varma and Richard Simon.
\newblock Bias in error estimation when using cross-validation for model
  selection.
\newblock \emph{BMC Bioinformatics}, 7:\penalty0 91, 2006.
\newblock \doi{10.1186/1471-2105-7-91}.

\end{thebibliography}

\end{document}